\RequirePackage{tikz}
\documentclass[sn-mathphys]{sn-jnl}

\usepackage[T1]{fontenc}
\usepackage[utf8]{inputenc}

\usepackage{subcaption}

\usepackage{booktabs} 

\usepackage{listings}
\usetikzlibrary{matrix,backgrounds,shapes.multipart,positioning,calc,chains,arrows,arrows.meta,decorations.pathmorphing,tikzmark,shadows,trees,mindmap,backgrounds}
 
\usepackage{pgfplots}
\pgfplotsset{%
    ,compat=1.12
    ,every axis x label/.style={at={(current axis.right of origin)},anchor=north west}
    ,every axis y label/.style={at={(current axis.above origin)},anchor=north east}
    }

\usepackage{etoolbox}

\jyear{2022}%

\theoremstyle{thmstyleone}%

\theoremstyle{thmstyletwo}%

\theoremstyle{thmstylethree}%
\title[Alleviating Overfitting in TIR SR with MOO]{Alleviating Overfitting in Transformation-Interaction-Rational Symbolic Regression with Multi-Objective Optimization}

\author*[1]{\fnm{Fabrício} \spfx{Olivetti} \sur{de França}} \email{folivetti@ufabc.edu.br}

\affil*[1]{\orgdiv{Center for Mathematics, Computation and Cognition}, \orgname{Federal University of ABC}, \orgaddress{\street{Av. dos Estados}, \city{Santo Andre}, \postcode{09210580}, \state{SP}, \country{Brazil}}}

\abstract{
    The Transformation-Interaction-Rational is a representation for symbolic regression that limits the search space of functions to the ratio of two nonlinear functions each one defined as the linear regression of transformed variables. This representation has the main objective to bias the search towards simpler expressions while keeping the approximation power of standard approaches.
   The performance of using Genetic Programming with this representation was substantially better than with its predecessor (Interaction-Transformation) and ranked close to the state-of-the-art on a contemporary Symbolic Regression benchmark. On a closer look at these results, we observed that the performance could be further improved with an additional selective pressure for smaller expressions when the dataset contains just a few data points. The introduction of a penalization term applied to the fitness measure improved the results on these smaller datasets. One problem with this approach is that it introduces two additional hyperparameters: i) a criteria to when the penalization should be activated and, ii) the amount of penalization to the fitness function.
   One possible solution to alleviate this additional burden of correctly setting these hyperparameters is to pose the search as a multi-objective optimization problem by minimizing the approximation error and the expression size. The main idea is that the selective pressure of finding non-dominating solutions will return the simplest model for each particular approximation error in the pareto front.
   In this paper, we extend Transformation-Interaction-Rational to support multi-objective optimization, specifically the NSGA-II algorithm, and apply that to the same benchmark. A detailed analysis of the results show that the use of multi-objective optimization benefits the overall performance on a subset of the benchmarks while keeping the results similar to the single-objective approach on the remainder of the datasets. Specifically to the small datasets, we observe a small (and statistically insignificant) improvement of the results suggesting that further strategies must be explored.
}


\begin{document}

\maketitle

\section{Introduction}~\label{sec:introduction}
 
Regression analysis studies the relationship between independent and dependent variables~\cite{kass1990nonlinear, harrell2017regression} with applications in a broad set of fields like economics, public health, engineering~\cite{gelman2020regression}.

This kind of analysis is traditionally performed using a pre-defined parametric models that are already well understood and can adapt to the data by adjusting their numerical parameters~\cite{kass1990nonlinear,harrell2017regression,gelman2020regression}.
These techniques have the advantage of having an extensive set of tools created to find an optimal parameters set. On the other hand, working with a fixed function can limit the possible shapes the regression model can fit, limiting their extrapolation capabilities~\cite{kronberger2022shape,haider2022comparing}. 

As an alternative approach, Symbolic Regression (SR)~\cite{koza1992genetic} searches for a function altogether with the adjusted parameters that best fits a studied phenomena. This has the advantages that no prior assumptions are required (e.g., linearity, homoscedasticity, etc.) as well as it is not limited to a certain class of shapes. The evolutionary approach called \emph{Genetic Programming} (GP)~\cite{koza1992genetic, koza1994genetic, poli2008field} is often employed to solve this particular problem.

In the standard approach, the representation of a solution in GP is an expression tree data structure that is only constrained by the set of symbols determined by the user. The general expectation is that the evolutionary process will return an accurate regression model. In practice, though, this not always works as expected since the search space can be hard to navigate given that a small change in the expression can propagate to a very different prediction behavior. 

An alternative representation was proposed in~\cite{SymTree,LightweightSymbolicRegressionWiththeIT} constraining the search space to a specific pattern composed of an affine transformation of non-linear transformed features. These features are a composition of a polynomial, representing the interactions between the original variables, and the application of an unary function, called transformation (also known as basis function in regression analysis). This representation, called Interaction-Transformation, presented competitive results when compared to the current approaches from the literature while guaranteeing the absence of certain complicated constructs such as nonlinear function chaining and nonlinear parameters, which are harder to interpret. Additionally, given that the model is linear w.r.t. the parameters, the numerical values can be determined using an efficient ordinary least squares solver. In~\cite{ 2021srbench} this algorithm was tested on an extensive benchmark revealing that, with datasets with certain particularities, it did not present a competitive performance.

In~\cite{tir} the authors extended the IT representation as the rational of two IT expressions, with the representation and search algorithm called Transformation-Interaction-Rational (TIR). In this same paper they noticed a significant increase in rank when compared to its predecessor but, they also noticed that when the dataset was small, this technique (and possibly others) had a tendency to overfit. As such, they proposed a penalty function applied to the fitness whenever the algorithm detected a small dataset determined by some \textit{ad-hoc} rules. The introduction of the penalty function and rules increase the number of hyperparameter that must be determined before executing the algorithm or to be included in the grid search. One possibility mentioned by the authors was to modify the search algorithm to handle multiple objectives, the maximization of the accuracy score and the minimization of a complexity score, such that the selective pressure imposed by the dominance criteria would favor the selection of the simplest models for a certain range of accuracy score.

This paper extends the results analysis in~\cite{tir} to investigate the benefits of using multi-objective optimization (MOO) to add selective pressure favoring simpler expressions. The main objective is to understand whether MOO can replace the previously proposed penalized fitness, removing the need of having additional hyperparameters. As a secondary objective, we will also analyse the impact of MOO in the different subsets of benchmarks. For this purpose, we adapted the current framework to support Multi-Objective Optimization, more specifically the Non-dominating Sorting Genetic Algorithm (NSGA-II), and we apply this search algorithm with TIR in order to optimize for accuracy and the number of nodes of the expression tree, as a measure of simplicity. 

The remainder of this paper is organized as follows. In Section~\ref{sec:gp} we briefly explain the overfitting problem and report a literature review of how to alleviate this problem in symbolic regression. In Section~\ref{sec:tir}, we describe the TIR representation, the current implementation, and explain the Non-dominating Sorting Genetic Algorithm and its implementation in TIR framework. The experimentation method is described in Section~\ref{sec:method} followed by the results and discussions in Section~\ref{sec:results}. Finally, in Section~\ref{sec:conclusion}, we conclude this paper with a summary of the results and discussing some future steps.

\section{Overfitting in Symbolic Regression}\label{sec:gp} 

When performing regression analysis, the main desiderata is to use the most parsimonious model that captures the studied relationship among the features of interest~\cite{gelman2020regression, hawkins2004problem}. In other words, we want to apply the Occam's Razor principle and avoid adding more degree-of-freedoms or flexibility to the regression model than needed. For example, if two of the measured features are already sufficient for a good fit and if a linear model is enough, we should not include other features or fit a polynomial regression. Doing so can add to the cost of acquiring new measurements and it can capture the randomness of a noisy component of the measurement process. 
In machine learning in general, a model is overfitted whenever it fits the data (almost) perfectly without retaining a generalization ability for predicting future data. Particularly to regression this is often associated with the complexity of the model (i.e., number of parameters, degree of polynomial, etc.) and a lack of training samples.
More formally, whenever the returned model $f$ has a smaller error in the training samples than another model $f'$, but $f'$ has a smaller error when measured over the entire distribution of samples, it is said that $f$ overfits the training data~\cite{learning1997tom}. 
Whenever possible, overfitting is avoided by a careful examination of the nature of the data and using prior knowledge about the studied phenomena to select only the relevant features and make a conscious choice about the parametric model.

But, in some situations, a more detailed analysis may not be possible and, thus, generic approaches to alleviate such problem may be required. In regression analysis in general, we can apply a regularization term to the objective-function of the fitting process such as to stimulate a feature selection by setting the corresponding coefficients to zero. Another possibility, when considering more than one alternative models, is to use a cross-validation approach to select the one that generalizes better with samples outside the training data.

Particularly in SR, there is an additional degree-of-freedom to the regression process as the algorithm can choose any model among the search space of mathematical functions limited only on its primitives set and expression size limits. The size of hypothesis space is also linked to overfitting~\cite{ng1997preventing,cavaretta1999data}. 

In \cite{paris2003exploring}, the authors verified the presence of overfitting in GP under a symbolic regression and classification tasks. The authors used the GP as defined in~\cite{GPOnTheProgrammingOfComputersByMeansOfNaturalSelection}, GP with sizefair crossover, proposed in \cite{langdon2000size}, and basic GP in a boosting framework. In summary, they noticed the presence of overfitting in their experiments and noted that increasing the population size may have a negative effect. In their experiments, limiting the maximum depth of the trees had a positive effect on avoiding or alleviating the overfitting. Among the three tested approaches, the boosting GP was the most successful, noticeably reducing the effect of the noise.

A simple approach to stimulate simpler models, is the inclusion of a penalty coefficient into the fitness function~\cite{hastie2009elements, kronberger2011overfitting}. With this option the (maximization) fitness function becomes:

\begin{equation}
    f'(x) = f(x) - c \cdot l(x), 
\end{equation}
where $c$ is the penalty coefficient and $l(x)$ is a parsimony measure of the solution $x$. In SR this is usually the number of nodes of a tree representation of the model.

This approach has the negative aspect that, depending on the choice of the function $l$, the fitness function may be too sensitive to the parameter $c$, requiring an additional cost for tuning it. The optimal value for this hyperparameter may be different for different data sets and it can change for different values of the other hyperparameters common to GP. One solution to this problem is to make this coefficient dynamic or adaptive. For example, in~\cite{kronberger2011overfitting} the authors tested a dynamic coefficient proposed in~\cite{poli2008covariant} using the covariance between the expressions length and fitness and the variance of the lengths. In \cite{kronberger2011overfitting} the author further adapted this idea to only apply the penalized fitness when it detected overfitting. The authors have not observed significant difference among the tested variations (without penalization, with penalization, with penalization and overfitting detection) and noted that a more accurate measure of parsimony rather then expression length may return better results.


In~\cite{vanneschi2010measuring}, the authors proposed three measures to quantify bloat, overfitting, and functional complexity in SR. Particularly to overfitting, their measure consists in comparing the fitness of a training set to a validation set. If the validation fitness is better than the training fitness or the validation fitness is better than the all time validation fitness (from all previous generations), than there is no overfitting. Otherwise, it calculates the difference between the absolute difference of the validation and training fitness and the all time validation and training fitness. They point it out the limitation that this technique is sensitive to the choice of training and test data and suggests the use of a cross-validation approach.

In~\cite{chen2016improving} the authors proposed the use of Structural risk minimisation to estimate the generalization error of GP models. This is done by incorporating the confidence interval of the risk using an approximation of the calculation of the VC-dimension for a particular SR model. The results show a significant generalization error when compared to a standard GP approach and a GP with bias-variance decomposition.

Another strategy, recently proposed in~\cite{bomarito2022bayesian}, combines the Probabilistic Crowding selection scheme with a fitness measured by the normalized log-likelihood of the model composing them as a Bayesian model selection using the Fractional Bayes Factor. Their results reveal that this selection scheme has a positive effect on the generalization of the models, measured using a test set, and the model size.

Finally, in~\cite{kommenda2016evolving}, the authors tested the effect of applying a multi-objective approach to exert a selective pressure in the evolution process to favor more parsimonious expressions. The main idea was to change the objective-function to find a Pareto front of solutions that maximized the accuracy of the model and minimized a complexity measure of the expression. Even though the returned model was the most accurate expression in the front (and thus, the more complex), they argue that, if the population is Pareto optimal, this should be the simplest expression with that accuracy level.
The authors have tested different complexity measures and discretized the accuracy objective-function by rounding the $R^2$ score to three decimal places, thus a simpler model that is slightly less accurate than a more complex model will replace it in the Pareto front due to the dominance criteria. In their experiments, they noticed the benefit of this approach as it rids the practitioner of having to set a size limit for the expression or adjusting any penalization coefficient. Within the noiseless data, the multi-objective approach reached the same accuracy level as the single-objective but naturally limiting the expression size closer to the optimal setting obtained with the single-objective approach. With noisy data, the multi-objective approach was shown to be more competent to automatically determine the optimal expression length without any accuracy loss. 

Other studies have been performed on the effects of Multi-objective optimization in SR algorithms. Notably, in \cite{de2003multi} the authors noticed that using fitness and tree size as the objectives, the population has a tendency to converge to a larger set of smaller than average models leading to premature convergence. This draws the attention that when fighting bloat with MOO a diversity control mechanism should be enforce to alleviate this effect. In \cite{smits2005pareto} the authors studied the benefits of using MOO with GP in SR task. They found that MOO sped up the convergence of the algorithm, as it dealt with smaller expressions on average, and allowed the user to pick a model from a set of different trade-offs.

In \cite{burlacu2019parsimony}, the authors tested a combination of different parsimony measures (together with accuracy measure) in a multi-objective GP for SR showing that using a combination of more than two objectives can be benefical to generate a better pareto front (measured using hypervolume) and accuracy (measured with $R^2$). The most beneficial objectives were the accuracy, visitation length, and diversity. In some recent studies~\cite{haider2022comparing,ShapeConstrainedSymbolicRegressionImprovingExtrapolationPriorKnowledge,kubalik2021multi}, the multi-objective GP is used to enforce that the generated models behave following certain properties. For example, in~\cite{haider2022comparing,ShapeConstrainedSymbolicRegressionImprovingExtrapolationPriorKnowledge} the authors use the term \emph{shape-constraint} to define the constraints on the shape of the generated function such as monotonocity, concavity, etc.

In~\cite{tir} the overfitting problem was noticed in a subset of a comprehensive SR benchmark, called SRBench~\cite{2021srbench}. Specifically to the algorithm Transformation-Interaction-Rational (TIR), the author noted that in some seeds on the smaller datasets, the hyperparameter tuning stage sometimes returned a perfect training score ($R^2 = 1$) but with negative test scores, indicating an overfitting. The author alleviated this problem by introducing a penalty term to the fitness function, specifically to these smaller datasets (similar to \cite{kronberger2011overfitting}), which improved the overall results of the benchmark for the studied algorithm. In the next section, we will detail this algorithm, the adaptations to handle smaller datasets, and the adaptations proposed in this paper.

\section{Transformation-Interaction-Rational}\label{sec:tir}

The \emph{Interaction-Transformation} representation (IT)~\cite{SymTree} describes regression models as an affine transformation of non-linear functions applied to the product of interactions of the variables. The main idea comes from the observation that many engineering and physics equations can be described using this pattern.

Considering a tabular data set where each sampled point has $d$ variables  $\mathbf{x} = (x_{1}, x_{2}, \cdots, x_{d})$. The regression model following the Interaction-Transformation (IT) representation is described as:

\begin{equation} \label{eq: ExprIT}
    f_{\operatorname{IT}}(\mathbf{x, w}) = w_0 + \sum_{j = 1}^{m}{w_{j} \cdot (f_j \circ r_j) (\mathbf{x})},
\end{equation}
representing a model with $m$ terms where $\mathbf{w} \in \mathbb{R}^{m+1}$ are the coefficients of the affine transformation, $f_j : \mathbb{R} \rightarrow \mathbb {R}$ is the $j$-th transformation function and $r_j : \mathbb{R}^d \rightarrow \mathbb {R}$ is the interaction function:

\begin{equation} \label{eq: term}
    r_j(\mathbf{x}) = \prod_{i = 1}^{d}{x_i^{k_{ij}}},
\end{equation}

\noindent where $k_{ij} \in \mathbb{Z}$ represents the exponents for each variable. Whenever we fix the values of every $f_j$ and $k_{ij}$, the expression becomes a linear model w.r.t. its numerical parameters, as such we can find their optimal values with the ordinary least squares method. This is convenient for the main search algorithm that only needs to find the optimal value of $m$, the values for the exponents and the functions of each term. 

This constrained representation limits the number of possible functions that are contained in the search space, eliminating some complex and, sometimes, undesirable constructs, but it can also leave some patterns often observed in science out of the search space. This was noticed in~\cite{aldeia2020parametric} where the authors verified that approximately half of the Feynman benchmarks~\cite{AIFeynman} could not be represented with IT.

To alleviate this issue, in~\cite{tir} the authors proposed an extension to this representation called Transformation-Interaction-Rational representation (TIR). In short, it simply combines two IT expressions as a rational function, similar to the rational polynomial regression model~\cite{taavitsainen2010ridge, taavitsainen2013rational, moghaddam2017statistical} and apply an invertible function to the resulting value such as:

\begin{equation*}
    f_{\operatorname{TIR}}(\mathbf{x, w_p, w_q}) = g\left(\frac{p(\mathbf{x, w_p})}{1 + q(\mathbf{x, w_q})}\right),
\end{equation*}
where $g : \mathbb{R} \rightarrow \mathbb{R}$ is an invertible  function, $p, q : \mathbb{R}^d \rightarrow \mathbb{R}$ are IT expressions exactly as defined in Eq.~\ref{eq: ExprIT} with $m_p > 0$ and $m_q \geq 0$ terms.

Notice that with this representation, the numerical parameters $\mathbf{w_p, w_q}$ of the model become non-linear. But, if we assume a noiseless training data $(x_{tr}, y_{tr})$, we can rearrange the expression as:

\begin{align*}
    y_{tr} &= g\left(\frac{p(\mathbf{x, w_p})}{1 + q(\mathbf{x, w_q})}\right) \\
    g^{-1}(y_{tr}) &= \frac{p(\mathbf{x, w_p})}{1 + q(\mathbf{x, w_q})} \\ 
    g^{-1}(y_{tr}) &= p(\mathbf{x, w_p}) - g^{-1}(y_{tr}) \cdot q(\mathbf{x, w_q}).
\end{align*}

By doing so, we can adjust the coefficients using the same fitting algorithm (OLS) of the transformed training target by applying the inverse of the transformation function $g$. For that to work, it is required that the function $g$ is invertible and that $y_{tr}$ is within the domain of the inverted function when $g^{-1}$ is partial. Additionally, it is important that $p$ contains at least one term to avoid the trivial solution where $q(\mathbf{x, w_q}) = -1$.
The adjusted model is then evaluated with a validation set containing at least some distinct samples from the training data to avoid keeping overfitted models in the population. This rearrangement of the expression works well even under the presence of noise, as studied in~\cite{de2022comparison}, and returns non discernible results from using a nonlinear optimization method, as reported in~\cite{de2023transformation}.



The main algorithm follows the same steps as genetic programming (GP) and the specific details of each step and operators are given in the following subsections. 

Related to this work, FFX~\cite{mcconaghy2011ffx} fits a LASSO regression model on a large set ($B$) of linear and nonlinear features composed of the original features, original features with three different exponents ($0.5, 1.0, 2.0$) and application of basis functions, and two-way interactions of the original features with the exponents. As in TIR, FFX also creates the set of features $B \cdot y$ to fit a rational polynomial regression model. As LASSO returns a path of regularization, leading to several models with different compromises, FFX returns the non-dominated set of these different models considering the goodness-of-fit and number of selected features.
The main advantage is that this is a deterministic and fast approach that returns multiple solutions with different compromises. On the other hand, it is limited to a few exponents and only two-way interaction as higher-order interactions may lead to an exponential growth with the dimension of the data. 

Another related work, in~\cite{giustolisi2006symbolic}~\footnote{Until the time of the publication the author was unaware of this work and thus is } the authors represent a solution as a matrix of exponents corresponding to a function form of a sum of product of the variables raised to a certain power (i.e., as the IT-expression without the transformation function). The expressions are created using a traditional GA evolving a flattened matrix of integers where each integer represent one of the candidate exponents are set by the user. The terms of the sum are combined with adjustable parameters as in this work. This representation can also be extended by pre-determining transformation functions for each term of the sum, also set by the user. As an example, given two variables, the choice of $2$ terms, and the exponents $[-1, 0, 1]$, the expression $f(x; \theta) = \theta_0 + \theta_1 x_1 x_2^{-1} + \theta_2 x_2$ would be represented as the vector $[3, 1, 2, 1]$ corresponding to the choices of exponent. The main difference of this approach with TIR is that TIR uses a variable-sized tree representation and automatically determine the transformation function.

\subsection{Initial Population}

Each solution is represented as a triple $(g, p, q)$ where $g$ is the invertible function, and $p, q$ are IT-expressions represented as a list of triples $[(w_i, f_i, [k_{ij}])]$. 
The initial population is composed of random solutions limited by a user defined budget of the number of tokens. The procedure starts by first choosing a random invertible function $g$ from a set $G$ of invertible functions, then generating a random non-null IT expression $p$ and, finally, an IT expression $q$. To generate an IT expression we repeatedly create random terms (see Eq.\ref{eq: term}) until either this procedure returns an empty term or we do not have enough budget. 

Each term is generated iteratively choosing a variable without replacement from the set of variables or stopping the procedure with a probability of $1/(d+1)$ where $d$ is the number of unchosen variables.


\subsection{Crossover}

The crossover procedure will choose two parents through tournament selection to take part of the process. In the first step, the procedure draws a random point of the first parent to make the recombination. If this point is located at the transformation function $g$, it will generate a child with $g, p$ taken from the first parent and $q$ from the second (Fig.~\ref{fig:cx1}). If the point is located at the IT expression $p$, it will create a child with $g, q$ from the first parent and a new $p$ as the recombination of both parents (Fig.~\ref{fig:cx2}). Likewise, if the point is at $q$ the child will have $g, p$ from the first parent and a recombined $q$ (Fig.~\ref{fig:cx3}).

\begin{figure}[t!]
    \centering\footnotesize
\begin{tikzpicture}[ampersand replacement=\&,font=\ttfamily,
array/.style={matrix of nodes,nodes={draw, minimum size=6mm, anchor=center},column sep=-\pgflinewidth, row sep=0.8mm,text depth=.5ex,text height=1ex,text width=8em}, nodes in empty cells]

\matrix[array] (array) {
 $g$ \& $p$ \& $q$ \\
 log \& $\tanh{(x_1^2 x_2)}$ \& $\sin{(x_1^3)} + x_0^2$ \\
 id \& $x_1 x_2^{-1} + \exp{(x_2)}$ \& $x_1 x_3^2$ \\
 log \& $\tanh{(x_1^2 x_2)}$ \& $x_1 x_3^2$ \\
                       };

\begin{scope}[on background layer]
\fill[gray!30] (array-1-1.north west) rectangle (array-1-3.south east);
\fill[red!30] (array-2-1.north west) rectangle (array-2-1.south east);
\fill[green!30] (array-2-2.north west) rectangle (array-2-3.south east);
\fill[blue!50] (array-3-1.north west) rectangle (array-3-3.south east);
\fill[green!30] (array-4-1.north west) rectangle (array-4-2.south east);
\fill[blue!50] (array-4-3.north west) rectangle (array-4-3.south east);
\end{scope}
\end{tikzpicture}
\caption{Crossover when the crossing point of the first parent is located at the root node ($g$). $g, p$ is taken from the first parent and $q$ from the second parent.}
\label{fig:cx1}
\end{figure}

\begin{figure}[t!]
    \centering\footnotesize
\begin{tikzpicture}[ampersand replacement=\&,font=\ttfamily,
array/.style={matrix of nodes,nodes={draw, minimum size=6mm, anchor=center},column sep=-\pgflinewidth, row sep=0.8mm,text depth=.5ex,text height=1ex,text width=8em}, nodes in empty cells,fill fraction/.style n args={2}{path picture={
 \fill[#1] (path picture bounding box.south west) rectangle
 ($(path picture bounding box.north west)!#2!(path picture bounding box.north
 east)$);}}]

\matrix[array] (array) {
 $g$ \& $p$ \& $q$ \\
 log \& $\tanh{(x_1^2 x_2)}$ \& $\sin{(x_1^3)} + x_0^2$ \\
 id \& $x_1 x_2^{-1} + \exp{(x_2)}$ \& $x_1 x_3^2$ \\
 log \& $\tanh{(x_1^2 x_2^{-1})}$ \& $\sin{(x_1^3)} + x_0^2$ \\
                       };

\begin{scope}[on background layer]
\fill[gray!30] (array-1-1.north west) rectangle (array-1-3.south east);
\fill[red!30] (array-2-2.north west) rectangle (array-2-2.south east);
\fill[green!30] (array-2-1.north west) rectangle (array-2-1.south east);
\fill[green!30] (array-2-3.north west) rectangle (array-2-3.south east);
\fill[blue!50] (array-3-1.north west) rectangle (array-3-3.south east);
\fill[green!30] (array-4-1.north west) rectangle (array-4-1.south east);
\fill[green!30, fill fraction={blue!50}{0.5}] (array-4-2.north west) rectangle (array-4-2.south east);
\fill[green!30] (array-4-3.north west) rectangle (array-4-3.south east);
\end{scope}
\end{tikzpicture}
\caption{Crossover when the crossing point of the first parent is located at the $p$ expression. In this case $g$ and $q$ are inherited from the first parent and $p$ is a mix of both parents.}
\label{fig:cx2}
\end{figure}

\begin{figure}[t!]
    \centering\footnotesize
\begin{tikzpicture}[ampersand replacement=\&,font=\ttfamily,
array/.style={matrix of nodes,nodes={draw, minimum size=6mm, anchor=center},column sep=-\pgflinewidth, row sep=0.8mm,text depth=.5ex,text height=1ex,text width=8em}, nodes in empty cells, fill fraction/.style n args={2}{path picture={
 \fill[#1] (path picture bounding box.south west) rectangle
 ($(path picture bounding box.north west)!#2!(path picture bounding box.north
 east)$);}}]

\matrix[array] (array) {
 $g$ \& $p$ \& $q$ \\
 log \& $\tanh{(x_1^2 x_2)}$ \& $\sin{(x_1^3)} + x_0^2$ \\
 id \& $x_1 x_2^{-1} + \exp{(x_2)}$ \& $x_1 x_3^2$ \\
 log \& $\tanh{(x_1^2 x_2)}$ \& $\sin{(x_1^3)} + x_1 x_3^2$ \\
                       };

\begin{scope}[on background layer]
\fill[gray!30] (array-1-1.north west) rectangle (array-1-3.south east);
\fill[red!30] (array-2-3.north west) rectangle (array-2-3.south east);
\fill[green!30] (array-2-1.north west) rectangle (array-2-2.south east);
\fill[blue!50] (array-3-1.north west) rectangle (array-3-3.south east);
\fill[green!30] (array-4-1.north west) rectangle (array-4-2.south east);
\fill[green!30, fill fraction={blue!50}{0.5}] (array-4-3.north west) rectangle (array-4-3.south east);
\end{scope}

\end{tikzpicture}
\caption{Crossover when the crossing point of the first parent is located at the $q$ expression. In this case $g$ and $p$ are inherited from the first parent and $q$ is a mix of both parents.}
\label{fig:cx3}
\end{figure}

\subsection{Mutation}

The mutation procedure is a uniform random choice between the set of mutations \emph{insert-node, remove-node, change-var, change-exponent, change-function}. In the special cases where the expression is close to the budget limit or with a single term, the operators \emph{insert-node} and \emph{remove-node} are removed from this set, respectivelly.

\begin{itemize}
    \item \textbf{insert-node:} it will choose with $50\%$ chance to either randomly insert a new variable in a randomly chosen term or create a new term in either $p$ or $q$ following the initialization procedure.
    \item \textbf{remove-node:} similar to \emph{insert-node}, but it will remove the variable or term entirely.
    \item \textbf{change-var:} it will swap a random variable in the expression with a different one.
    \item \textbf{change-exponent:} it will randomly replace one exponent from the expression.
    \item \textbf{change-function:} it will change one of the transformation functions with a random choice.
\end{itemize}

\subsection{Penalized Fitness}

When analysing the performance of TIR under the extensive SRBench benchmark~\cite{2021srbench,tir}, the author noticed that on a subset of the datasets, the number of samples was too small allowing a perfect fit on the training set that translated into a bad performance in the test set used for evaluation.
In this same paper, the author proposed the use of a penalized fitness function:

\begin{equation}
    f'(x) = f(x) - c \cdot l(x),
\end{equation}
where $f(x)$ is the original maximization fitness function, $c$ is the penalization coefficient and $l(x)$ is the number of nodes in expression $x$.

In some pre-analysis, the author noticed that a penalization coefficient $c = 0.01$ helped to alleviate this issue but only if applied to the small datasets. To decide whether to apply this penalization, the author proposed three \emph{ad-hoc} heuristics:

\begin{itemize}
    \item \textbf{TIR-samples:} the dataset is small if the number of samples is less than or equal $100$.
    \item \textbf{TIR-dim:} the dataset is small if the number of features is less than or equal $6$.
    \item \textbf{TIR-points:} the dataset is small if the product of the number of samples and number of features is less than or equal $1000$.
\end{itemize}

In a detailed analysis of the results~\cite{de2023transformation}, it was noted that TIR-points was the most successful among the tested versions of TIR. Arguably, the decision heuristic and the penalization coefficient are additional hyperparameters to the algorithm and it should also be optimized within the cross-validation process during a focused regression analysis. The main reason to turn those into empirical values is the limit imposed by SRBench of allowing only six different hyperparameters configuration.

\subsection{Handling Overfitting with Multi-objective Optimization}

Novel to this work, we will replace the main evaluation and reproduction steps of the algorithm with a Multi-objective optimization (MOO) approach, specifically the Fast Non-dominated sorting algorithm (NSGA-II) and the Crowding Distance~\cite{deb2000fast}.

In the context of overfitting, the MOO approach can naturally exert a selective pressure to the population stimulating that the best model is the smallest model with the maximum accuracy. In a perfect situation that the Pareto optimal set is returned, this is true as given two solutions with the same accuracy, the smaller one would dominate the other. This idea is similar to what was explored in~\cite{kommenda2016evolving}. The main advantage of this approach is that it does not require additional hyperparameters and it will automatically adjust the maximum length of the expression. On the other hand, the calculation of the front adds a significant computational cost to the algorithm.
Particular to this work, the MOO variant of TIR (called TIRMOO) with the optimization of two objectives, $f_1$ as the maximization of $R^2$ score and $f_2$ as the minimization of the number of nodes in the expression tree. While the definition of \emph{simplicity} in SR is subjective and a topic of research on its own, the number of nodes serves as a proxy for simplicity. As TIR representation already alleviates one of the patterns commonly associated with functional complexity, nonlinear functions chaining, the length criteria may be enough to stimulate the search towards simpler solutions.
We will also test four different variants that are different in how we choose the returned solution, whether to use a penalized accuracy objective, and when to apply these heuristics:

\begin{itemize}
    \item \textbf{TIRMOO:} the plain TIR algorithm using MOO instead of single-objective, the returned model is the most accurate of the front.
    \item \textbf{TIRMOO-points:} same as the previous approach but replacing the accuracy objective with the penalized version applied only to datasets where the product of number of samples and the number of features are less than or equal $1000$.
    \item \textbf{TIRMOO-Select:} same as TIRMOO but selecting the smallest model of the front with accuracy within $95\%$ of the most accurate model.
    \item \textbf{TIR-Sel-points:} it returns the most accurate model of the front (as TIRMOO) but returns the same choice as TIRMOO-Select on datasets where the product of number of samples and the number of features are less than or equal $1000$ (same selection as TIRMOO-points).
\end{itemize}

\section{Experiments}\label{sec:method}

To assess the results of Non-Dominated Sorting as a reproduction mechanism of TIR, we executed the  symbolic regression benchmark \emph{SRBench}~\footnote{\url{https://cavalab.org/srbench/}} to evaluate and compare the single and multi objective version of TIR. This benchmark contains $122$ regression problems without any knowledge about the generating function. To make the comparisons fair, we follow the same methodology as described in the benchmark repository, executing each algorithm $10$ times with a fixed set of random seeds for every dataset. In each run, the benchmark splits the dataset with a ratio of $0.75/0.25$ for training and testing, subsampling the training data in case it contains more than $10,000$ samples. After that, it performs a halving grid search~\cite{jamieson2016non} with $5$-fold cross-validation to choose the optimal hyperparameters among a maximum of $6$ user defined choices. Finally, it executes the algorithm with the whole training data using the optimal hyperparameters to determine the final regression model. It evaluates this model in the test set and stores the approximation error information, running time, number of nodes, and the symbolic model.

SRBench calculates and store the mean squared error (mse), mean absolute error (mae), and coefficent of determination ($R^2$) measurements. The results of each dataset are summarized using the median to minimize the impact of outliers and they are reported with error plots sorted by the median of this summarization. In this work we will focus on the $R^2$ metric following the same analysis in~\cite{2021srbench}.

Currently the \emph{srbench} repository contains results of $21$ regression algorithms, $14$ of which are Symbolic Regression algorithms. From these SR algorithms, $10$ use an evolutionary algorithm as a search procedure. It is important to notice that $62$ of the $122$ datasets are variations of the Friedman benchmark with different numbers of variables, samples, and noise levels. As such, we will also report a separate analysis for the Friedman, non-Friedman sets, and the datasets selected by the \emph{points} heuristic as described in the previous section.


As all the results reported in~\cite{2021srbench} are publicly available at their repository, we only executed the experiments for TIR with the same random seeds so that the results remain comparable. We executed these experiments using Intel DevCloud instance with an Intel(R) Xeon(R) Gold 6128 CPU @ 3.40GHz using only a single thread. The original experiment was executed on an Intel(R) Xeon(R) CPU E5-2690 v4 @ 2.60GHz, so this may create a bias when comparing the execution time. The results and the processing scripts are available at \url{https://github.com/folivetti/tir/tree/main/papers/GPEM}.

\subsection{Hyperparameters}

The rules for this benchmark framework allows only the use of $6$ hyperparameters settings for the grid search procedure. Since the MOO approach demands a higher computational execution time, we decided to fix the hyperparameters based on the most frequently selected values from TIR in previous experiments (upper part of Table~\ref{tab:fixedpar}) and test two different settings for the $k_{ij}$ range (lower part of Table~\ref{tab:fixedpar}).

\begin{table}[t!]
    \centering
    \caption{Hyperparameters for the TIR algorithm used during the regression benchmark experiments.}
    \begin{tabular}{cc}
        \toprule 
        Parameter &  value \\ \midrule 
        Pop. size &  $1000$ \\ 
        Gens.     & $500$ \\ 
        Cross. prob. & $30\%$ \\ 
        Mut. prob & $70\%$ \\ 
        Transf. functions &  $[id,tanh, sin, cos, log, exp, sqrt]$ \\ 
        Invertible functions     & $[id,atan, tan, tanh, log, exp, sqrt]$ \\ 
        Error measure & $R^2$ \\ 
        budget & $max(5, min(15, \left\lfloor\frac{n\_samples}{10}\right\rfloor)$ \\ 
        \midrule
        $k_{ij}$ range (Eq.~\ref{eq: term}) & $\{(-5, 5), (0, 3) \}$ \\
        \bottomrule
    \end{tabular}
    \label{tab:fixedpar}
\end{table}

\section{Results}\label{sec:results}

In this section we will report the obtained results similarly to how it was done in~\cite{tir}, with error bars and critical difference diagrams. The error bars display the median of medians of the $R^2$ calculated over the test sets with a bar representing the confidence interval of $0.95$ estimated using a thousand bootstrap iterations using the percentile method. 

\begin{figure}[t]
    \centering
    \begin{subfigure}[b]{0.4\linewidth}
    \includegraphics[width=0.96\linewidth]{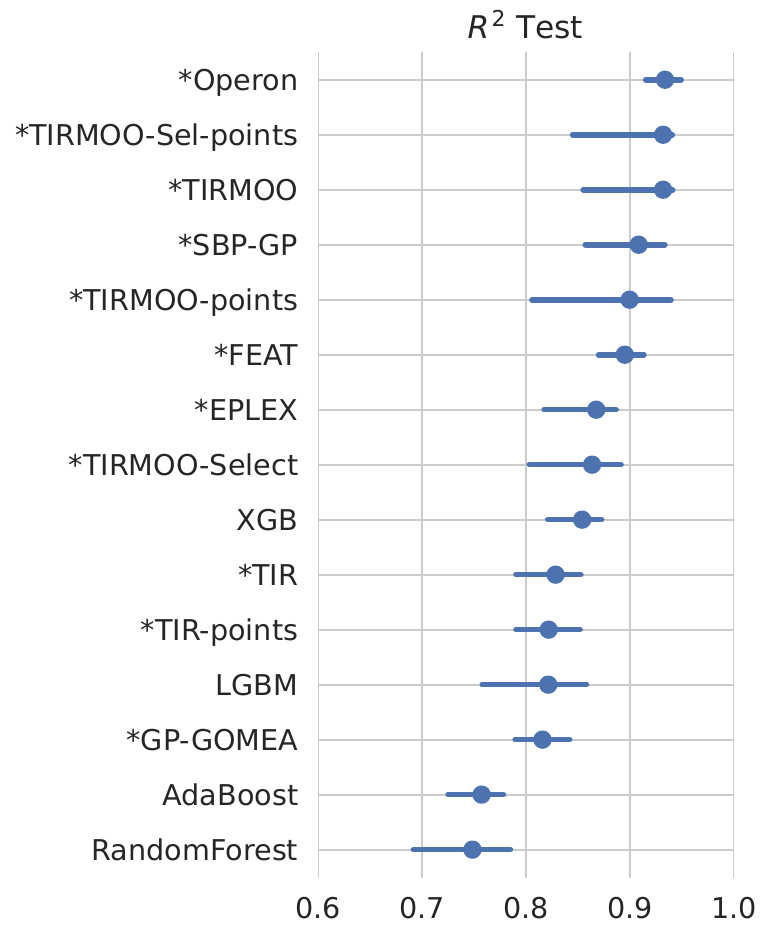}
    \caption{ }
    \label{fig:penalty}
    \end{subfigure}
    \begin{subfigure}[b]{0.4\linewidth}
    \includegraphics[width=0.96\linewidth]{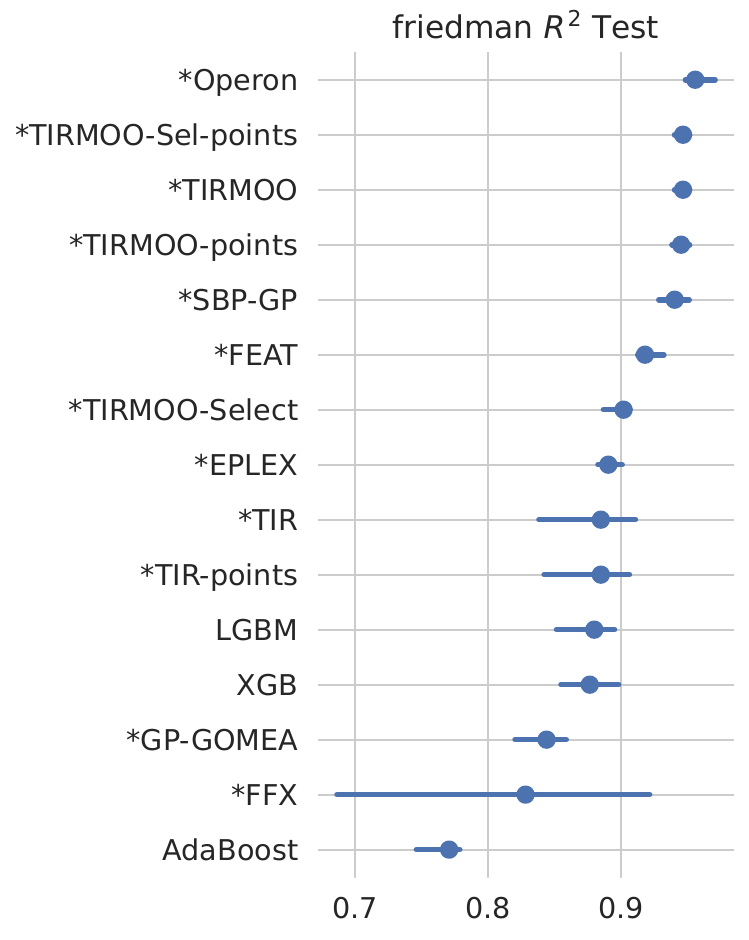}
    \caption{ }
    \label{fig:penalty_fri}
    \end{subfigure}
    \begin{subfigure}[b]{0.4\linewidth}
    \includegraphics[width=0.96\linewidth]{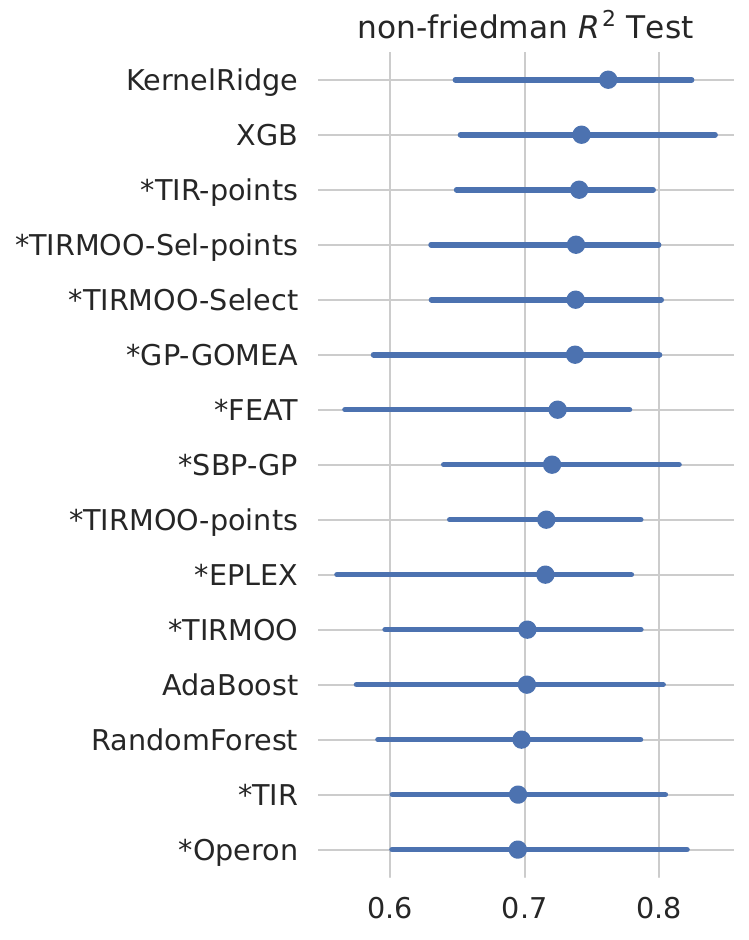}
    \caption{ }
    \label{fig:penalty_nonfri}
    \end{subfigure}
    \begin{subfigure}[b]{0.4\linewidth}
    \includegraphics[width=0.96\linewidth]{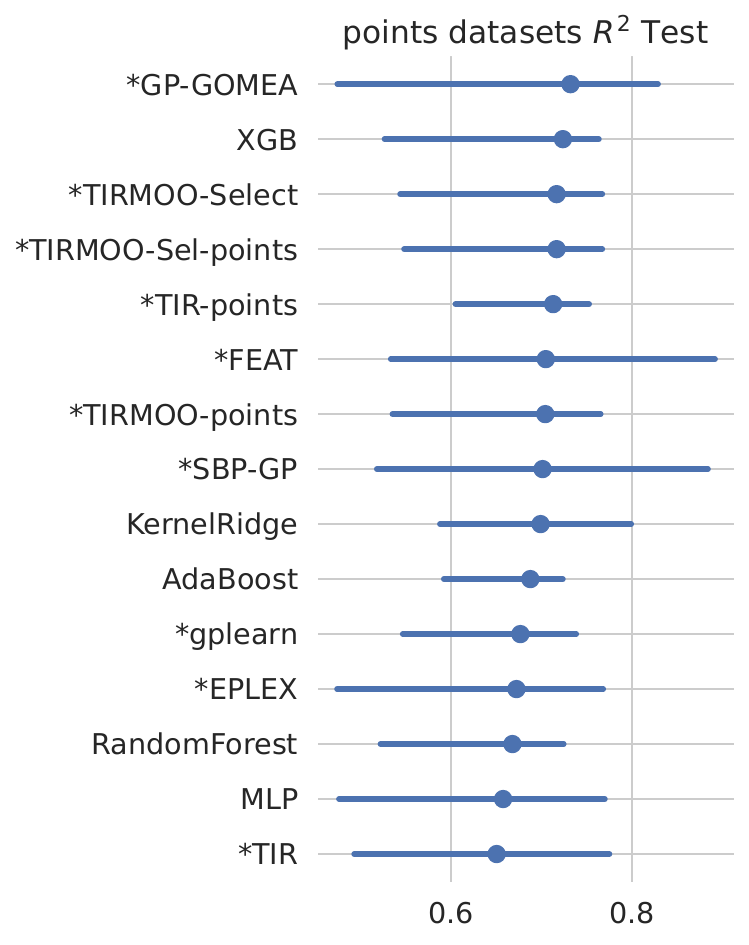}
    \caption{ }
    \label{fig:penalty_points}
    \end{subfigure}
    \caption{Top $15$ median of medians for (a) every data set, (b) Friedman data set, (c) non-Friedman datasets, and (d) datasets selected by points heuristic.}
\end{figure}

We can see in Figs.~\ref{fig:penalty}~to~\ref{fig:penalty_points} the error bars for the top $15$ algorithms considering every officially tested algorithm in \emph{srbench}, and every variation of TIR and TIRMOO. The first plot display the results for the entire benchmark, while the second and third plots show the rank for the Friedman and non-Friedman datasets, respectively. The last plot show the results only for the datasets as selected by the \emph{points} heuristic, this accounts for only $18$ datasets of the benchmark.

From the first plot we can see that TIRMOO and TIRMOO-Sel-points have very similar results with a median $R^2$ around $0.95$ with the error bar extending to $0.85$, this puts these two variants of TIRMOO together with Operon and SBP-GP in this range of accuracy. As reported in previous works, Operon~\cite{OperoncppAnEfficientGPFramework,kommenda2019parameter} stands out as the most accurate overall as it implements a standard tree-based Genetic Programming with a placement of optimized numerical parameters by means of nonlinear optimization.
Fig.~\ref{fig:penalty_fri} shows the results calculated only on the Friedman datasets. As reported in~\cite{de2023transformation}, this subset of datasets are determinant for a good rank in the overall scores. Operon ranks first with an $R^2$ larger than $0.95$ while TIRMOO, TIRMOO-Sel-points, and TIRMOO-points shows a median closer to this value. The $R^2$ score of most algorithms in the top-$15$ is between $0.85$ and $0.9$. As the error bar indicates, the top-$5$ performance is distinct from the remainder of the algorithms, unlike the first plot. Figs.~\ref{fig:penalty_nonfri}~and~\ref{fig:penalty_points} show the results for the non-Friedman datasets and the selection of smaller datasets (as per the \emph{points} heuristic). Even though Operon do not appear in these plots, its performance is also very close to these algorithms. The difference in results for the top-$15$ algorithms is negligible with a high intersection of the error bars.
In summary, most algorithms in the benchmark presents a similar behavior on the non-Friedman datasets, but the top-$5$ stands out in the Friedman datasets, elevating their positions in the rank.
Regarding the smaller datasets, the main difference is in the width of the error bars, showing that the TIR (and TIRMOO) variations with specific mechanisms to favor simpler expressions, displays a smaller error indicating less variability in the results. Other algorithms showing this behavior are non-symbolic methods like AdaBoost, Kernel Ridge and, the basic GP implemented in gplearn. Notice also the absence of plain TIRMOO in this last plot, indicating that the strategy of relying on the most accurate model in the MOO front is not enough to alleviate the problem with smaller datasets. As it turns out, TIRMOO-Sel-points seems to return the best compromise among all the tested variations. This indicates that TIR, and possibly SR in general, may benefit from a contextualized model selection approach that increases the preference for simpler model if the dataset is small and noisy.

To illustrate the overfitting behavior observed in these smallers datasets, Table~\ref{tab:overfit} shows the training and test $R^2$ of the worst and best solutions obtained by TIRMOO and TIRMOO-points in each one of these datasets, as selected using the test accuracy. We can see from this table, without a penalization factor, there are some solutions in which the model performs well in the training data, but with a negative $R^2$ in the test set. 

\begin{table}[t!]
\centering
\caption{Worst and best test accuracies obtained by TIRMOO and TIRMOO-points variations on the datasets where the number of samples multiplied by the number of features is smaller than $1000$. Negative $R^2$ are represented as <0 as some of these cases have a large magnitude.}
\footnotesize
\begin{tabular}{lllll|llll}
\hline 
& \multicolumn{4}{c|}{TIRMOO} & \multicolumn{4}{c}{TIRMOO-points} \\
\hline
& \multicolumn{2}{c}{worst test $R^2$} & \multicolumn{2}{c|}{best test $R^2$} & \multicolumn{2}{c}{worst test $R^2$} & \multicolumn{2}{c}{best test $R^2$}\\
\hline
dataset & train & test &  train & test &  train & test &  train & test \\
\hline
192\_vineyard                   & 0.80  & <0                                         & 0.80  & 0.62      & 0.80  & 0.04       & 0.75  & 0.69      \\
228\_elusage                    & 0.91  & <0                                          & 0.93  & 0.71      & 0.87  & 0.51       & 0.86  & 0.77      \\
485\_analcatdata\_vehicle       & 0.94  & <0                                          & 0.79  & 0.89      & 0.87  & <0      & 0.74  & 0.89      \\
1096\_FacultySalaries           & 1.0   & <0                                          & 0.99  & 0.98      & 1.0   & <0      & 0.89  & 0.91      \\
523\_analcatdata\_neavote       & 0.45  & 0.72                                           & 0.97  & 0.97      & 0.95  & 0.75       & 0.95  & 0.97      \\
663\_rabe\_266                  & 1.0   & 1.0                                            & 1.0   & 1.0       & 0.99  & 0.97       & 0.98  & 0.99      \\
687\_sleuth\_ex1605             & 0.88  & <0                                       & 0.88  & 0.50      & 0.75  & <0      & 0.78  & 0.42      \\
659\_sleuth\_ex1714             & 0.99  & <0                                         & 0.99  & 0.83      & 0.88  & 0.21       & 0.92  & 0.87      \\
678\_visualizing\_env. & 0.68  & <0 & 0.62  & 0.53      & 0.54  & <0  & 0.52  & 0.34      \\
611\_fri\_c3\_100\_5            & 0.93  & 0.52                                           & 0.94  & 0.93      & 0.90  & 0.65       & 0.87  & 0.87      \\
594\_fri\_c2\_100\_5            & 0.00  & <0                                          & 0.86  & 0.87      & 0.75  & <0      & 0.82  & 0.90      \\
624\_fri\_c0\_100\_5            & 0.89  & 0.66                                           & 0.88  & 0.84      & 0.87  & 0.61       & 0.87  & 0.86      \\
656\_fri\_c1\_100\_5            & 0.91  & 0.41                                           & 0.91  & 0.92      & 0.88  & 0.56       & 0.82  & 0.91      \\
210\_cloud                      & 0.95  & <0                                        & 0.91  & 0.94      & 0.87  & 0.49       & 0.81  & 0.95      \\
706\_sleuth\_case1202           & 0.88  & <0                                          & 0.84  & 0.68      & 0.84  & <0      & 0.82  & 0.76      \\
1089\_USCrime                   & 0.98  & <0                                          & 0.96  & 0.84      & 0.91  & 0.61       & 0.88  & 0.90      \\
712\_chscase\_geyser1           & 0.86  & <0                                         & 0.82  & 0.76      & 0.80  & 0.71       & 0.76  & 0.83     \\
\hline 
\end{tabular}
\label{tab:overfit}
\end{table}

\begin{figure}[t!]
    \centering
    \begin{subfigure}[b]{0.45\linewidth}
    \includegraphics[width=0.96\linewidth]{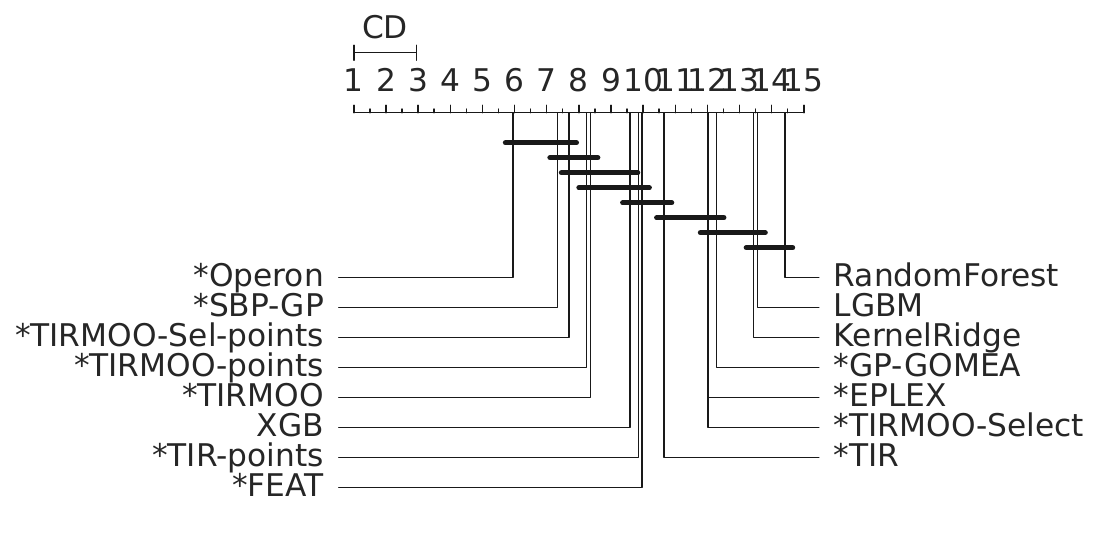}
    \caption{ }
    \label{fig:cd}
    \end{subfigure}
    \begin{subfigure}[b]{0.45\linewidth}
    \includegraphics[width=0.96\linewidth]{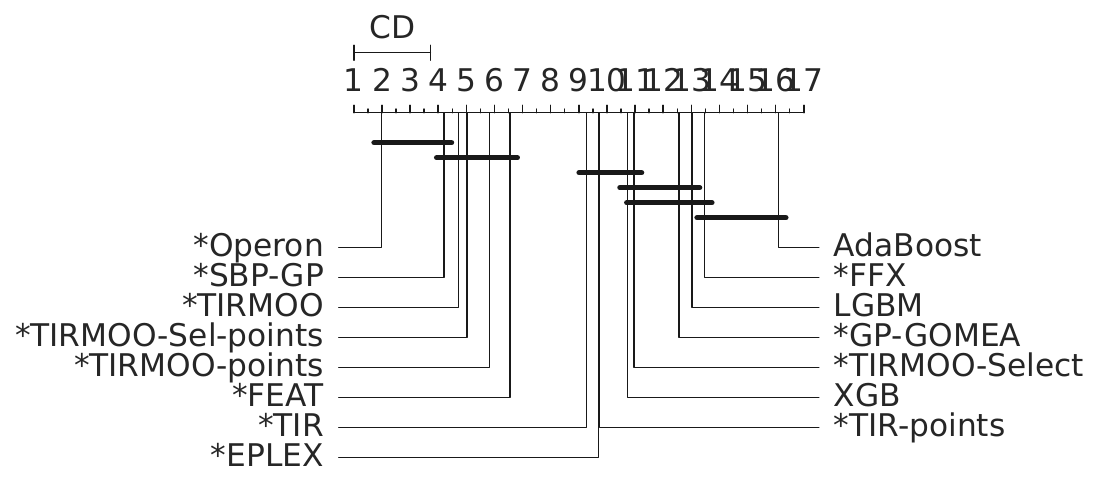}
    \caption{ }
    \label{fig:cd_fri}
    \end{subfigure}
    \begin{subfigure}[b]{0.45\linewidth}
    \includegraphics[width=0.96\linewidth]{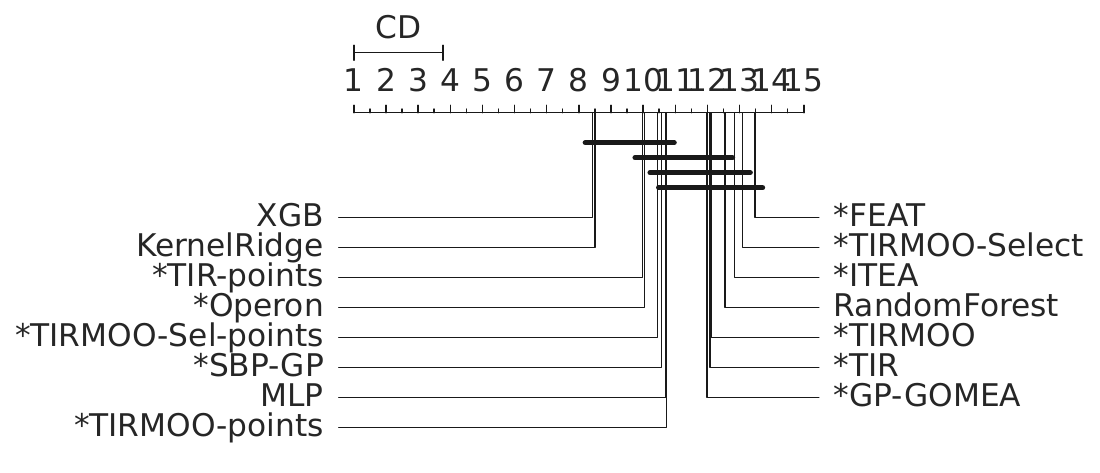}
    \caption{ }
    \label{fig:cd_nonfri}
    \end{subfigure}
    \begin{subfigure}[b]{0.45\linewidth}
    \includegraphics[width=0.96\linewidth]{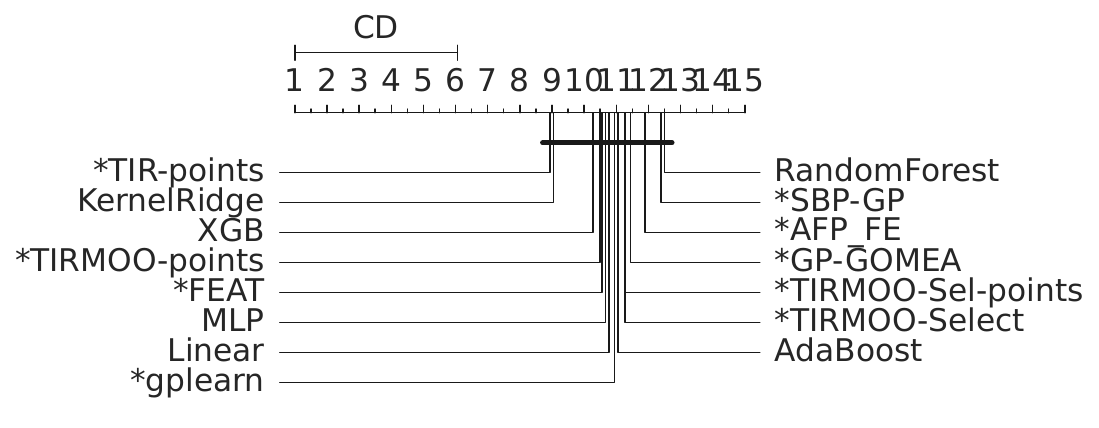}
    \caption{ }
    \label{fig:cd_points}
    \end{subfigure}
    \caption{Critical diagram of the top $15$ algorithms for (a) every data set, (b) Friedman data set, (c) non-Friedman datasets, and (d) datasets selected by points heuristic. These plots are computed using the Nemenyi test with $\alpha = 0.05$ calculated over the average rank.}
\end{figure}

The plots in Figs.~\ref{fig:cd}~to~\ref{fig:cd_points} show the critical difference diagrams using the Nemenyi test with $\alpha = 0.05$ as a post-hoc test to find the groups of algorithms with a significant difference to each other. This test is calculated using the average rank of each algorithm on each dataset using the median $R^2$ of the test set as the ranking criteria. Since these plots depict the average rank, they show a different view of the results.
For the overall results (Fig.~\ref{fig:cd}), we can see that Operon and SBP-GP do not present a statistically significant difference and TIRMOO, TIRMOO-Sel-points, TIRMOO-points does not present a statistically significance difference from SBP-GP. In Fig.~\ref{fig:cd_fri} we observe similar results for the Friedman datasets but, the average rank of the top-$5$ algorithms is between $2$ and $6$, whereas in the overal results it was between $6$ and $10$. This means that for this subset of the benchmark there are a small set of algorithms that stands out in comparison to the others. This is the opposite of what is observed in Figs.~\ref{fig:cd_nonfri}~and~\ref{fig:cd_points}, where the average ranks are between $9$ and $14$ and there is no significant difference between the different algorithms. The Kruskal-Wallis and Mood's median tests indicates that there is no significant difference among the top-$15$ algorithms in any of these selections. This corroborates with~\cite{de2023transformation} in which a more detailed analysis revealed that some noticeable difference is only observed on a specific selection of the datasets.

\begin{figure}[t!]
    \centering
    \begin{subfigure}[b]{0.45\linewidth}
    \includegraphics[width=0.96\linewidth]{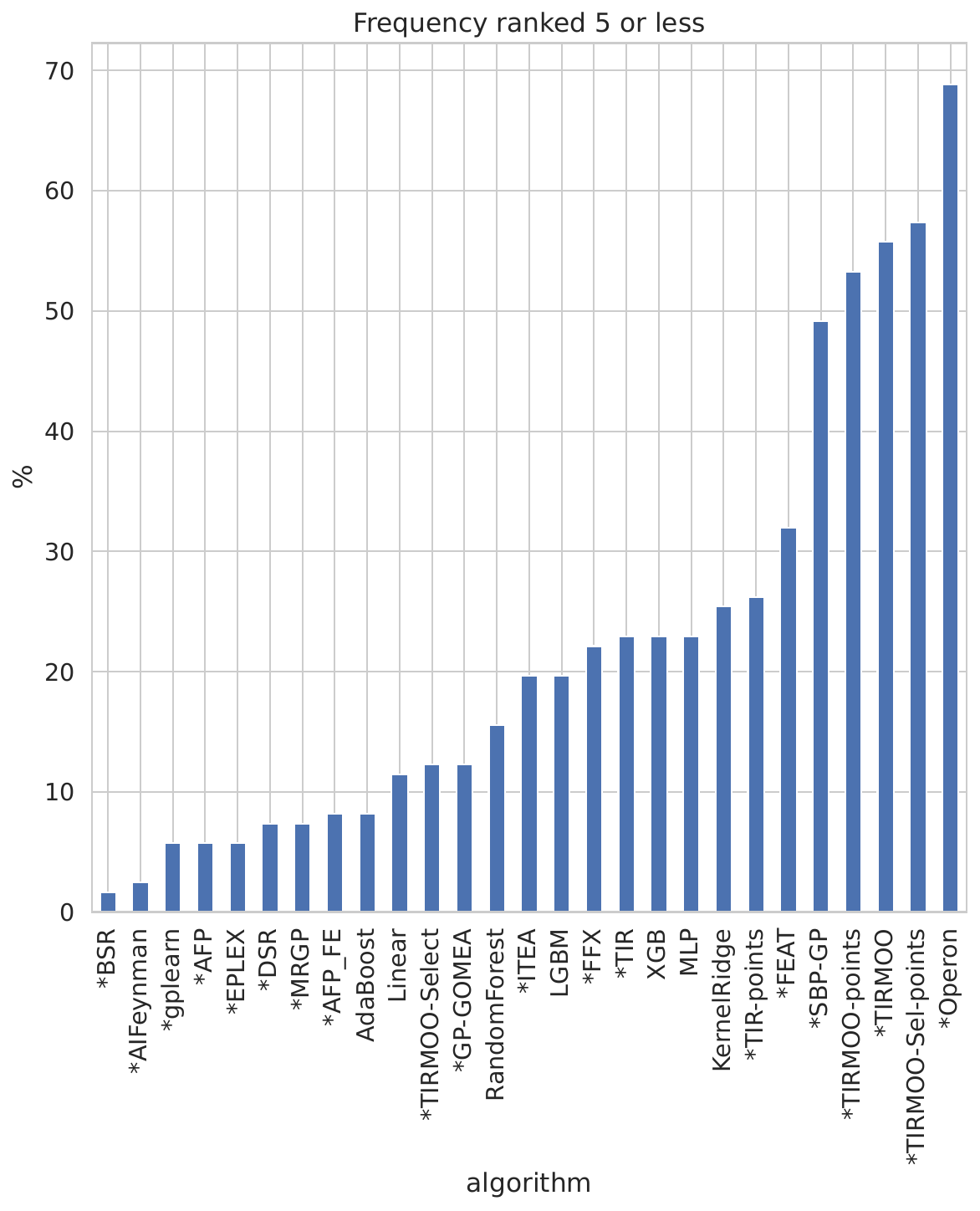}
    \caption{ }
    \label{fig:hist}
    \end{subfigure}
    \begin{subfigure}[b]{0.45\linewidth}
    \includegraphics[width=0.96\linewidth]{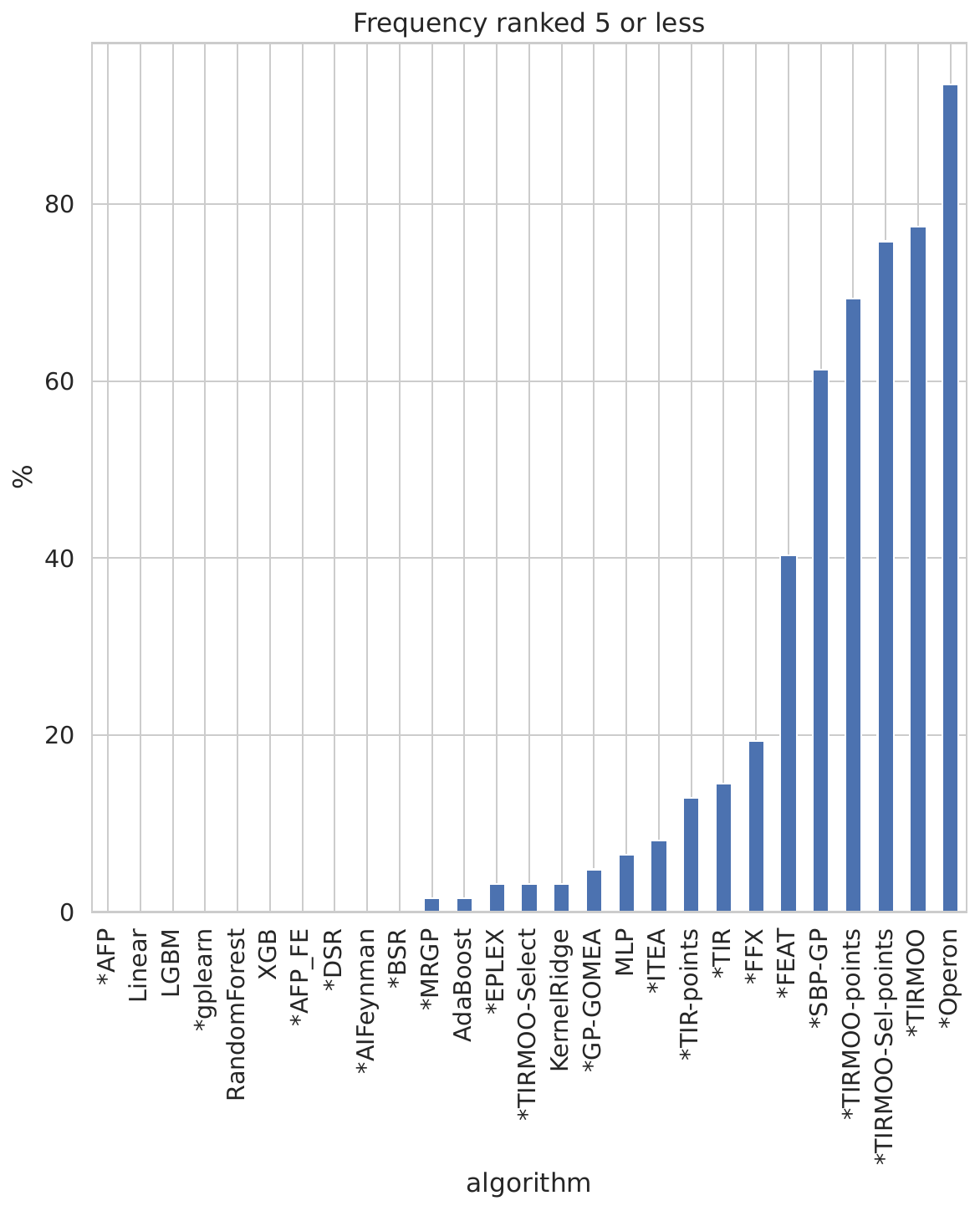}
    \caption{ }
    \label{fig:hist_fri}
    \end{subfigure}
    \begin{subfigure}[b]{0.45\linewidth}
    \includegraphics[width=0.96\linewidth]{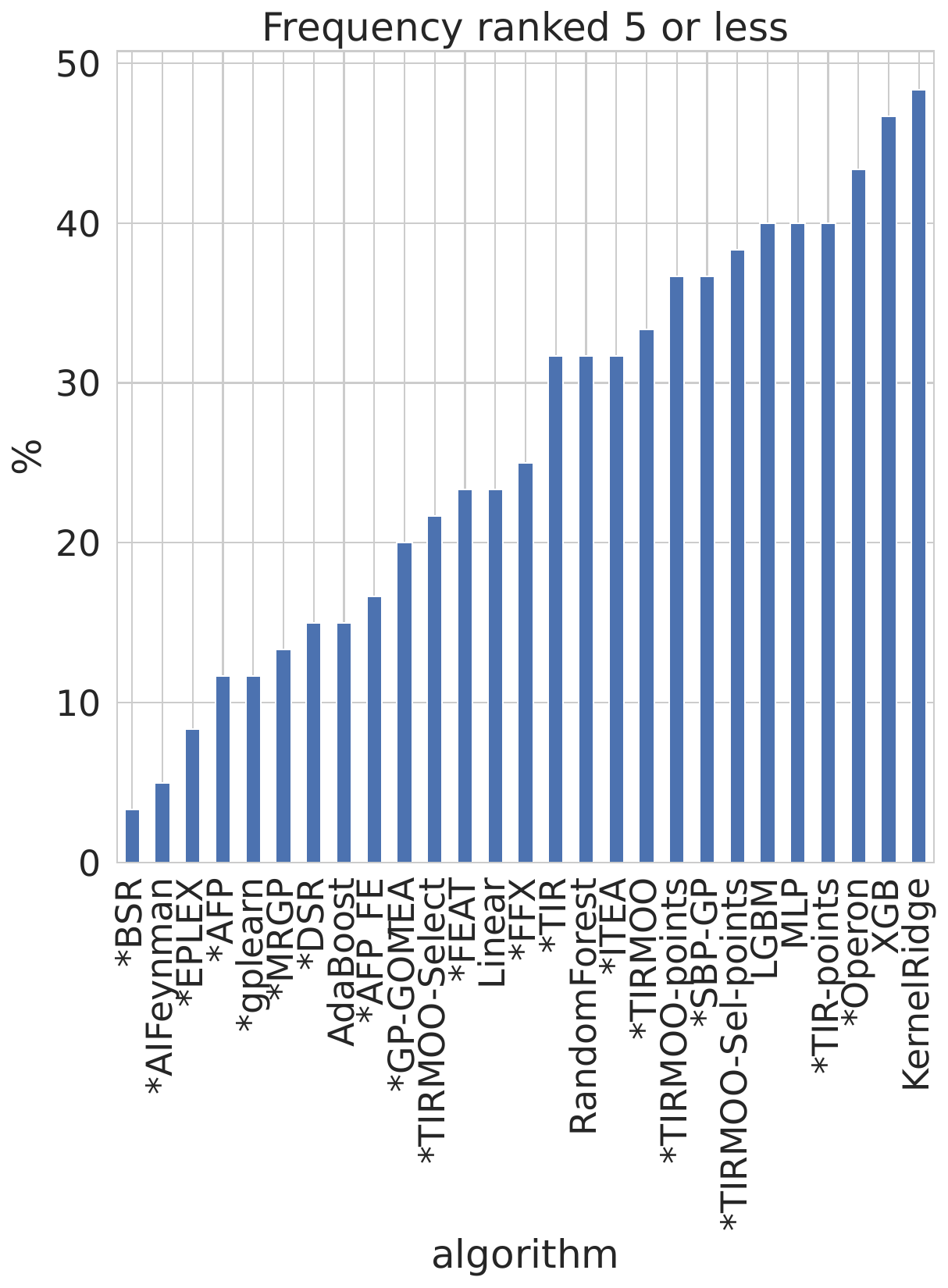}
    \caption{ }
    \label{fig:hist_nonfri}
    \end{subfigure}
    \begin{subfigure}[b]{0.45\linewidth}
    \includegraphics[width=0.96\linewidth]{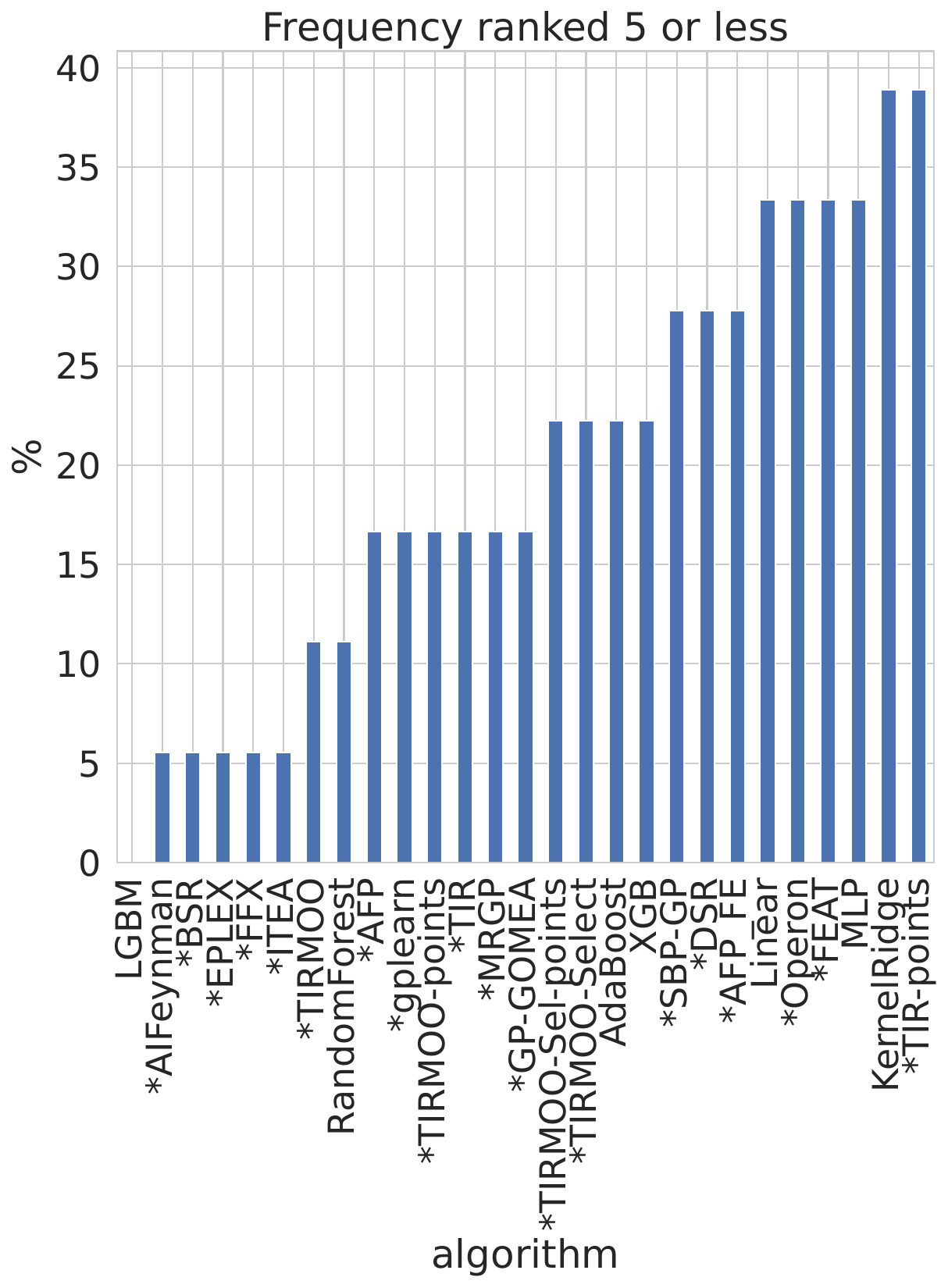}
    \caption{ }
    \label{fig:hist_points}
    \end{subfigure}
    \caption{Histogram of the counts of how many times each algorithm was ranked within the top-$5$ for (a) every data set, (b) Friedman data set, (c) non-Friedman datasets, and (d) datasets selected by points heuristic.}
\end{figure}

Finally, in Figs.~\ref{fig:hist}~to~\ref{fig:hist_points} we can see histograms that show the percentage of datasets (in each selection) in which each algorithm was ranked within the top-$5$. This histogram can reveal to what extent the best algorithms dominate the others. For the purpose of ranking we have rounded the $R^2$ score to the second decimal place before ranking the algorithms.
From these plots we can see that overall, Operon is very competent and reaches almost $70\%$ of the datasets within the top-$5$ followed by TIRMOO-Sel-points and TIRMOO accounting for about $55\%$ each. This is mostly because of the Friedman datasets in which these algorithms account for $88\%, 68\%, 65\%$, respectively. When looking at the non-Friedman and the small datasets, we see a degradation in TIRMOO performance as their best variations ranks within the top-$5$ in only $35\%$ and $25\%$ of these datasets, respectively. On the other hand, the single-objective version, TIR-points, was tied with Operon in the non-Friedman datasets ranked as top-$5$ in about $42\%$ of these datasets and outperformed all other methods in the smaller datasets, in $48\%$ (Operon was the top-$5$ in $35\%$ of this selection). Another highlight in these plots is the presence of FFX~\cite{mcconaghy2011ffx} algorithm that, as already mentioned, also fits rational polynomial models. This algorithm accounts for a very similar percentage as TIR in those plots but, both, with a smaller percentage than TIRMOO and its variants.

\subsection{Comparison between the different TIR variants}

\begin{figure}[t!]
    \centering
    \begin{subfigure}[b]{0.45\linewidth}
    \includegraphics[width=0.96\linewidth]{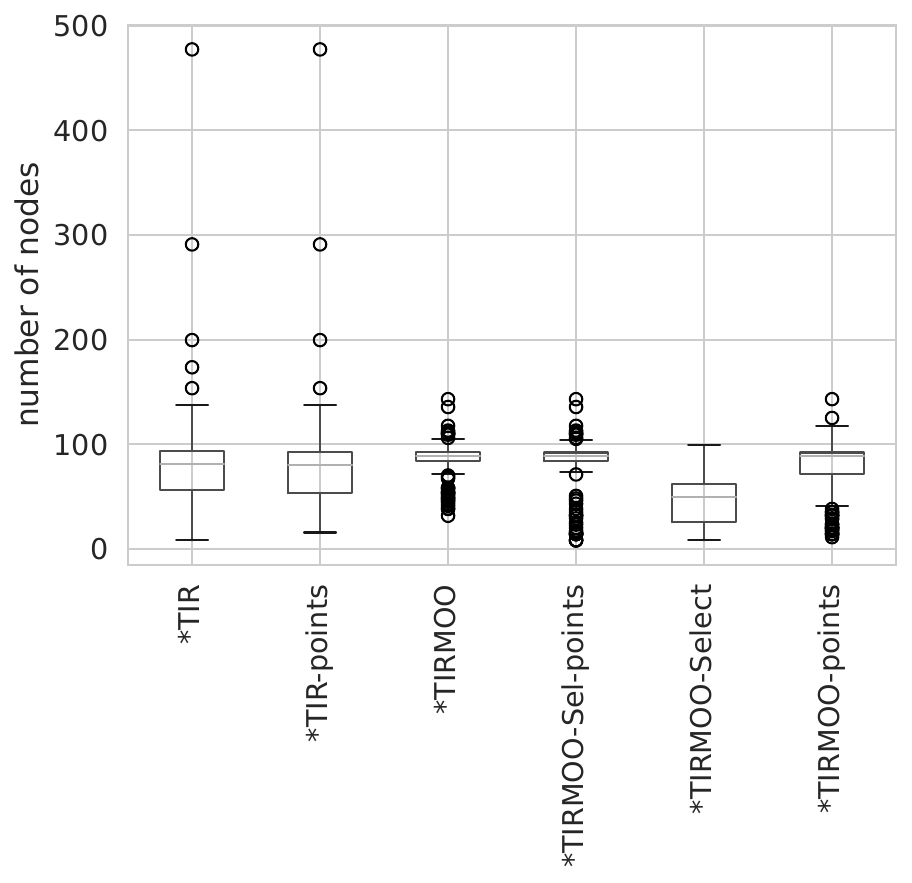}
    \caption{ }
    \label{fig:size}
    \end{subfigure}
    \begin{subfigure}[b]{0.45\linewidth}
    \includegraphics[width=0.96\linewidth]{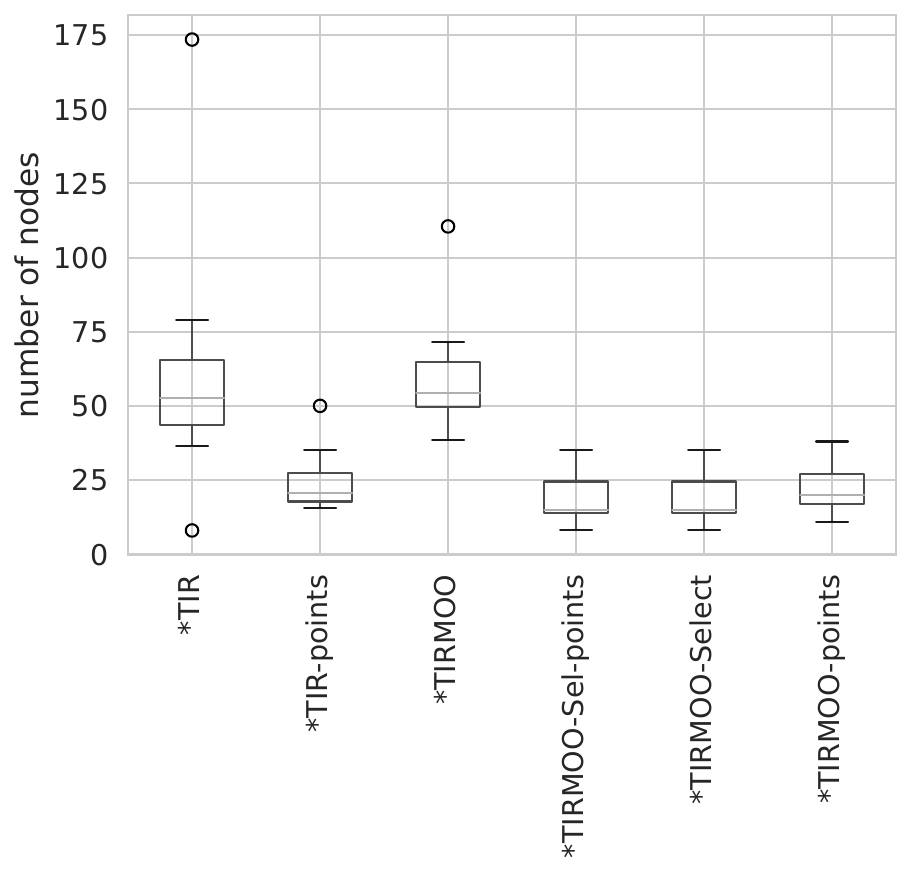}
    \caption{ }
    \label{fig:size-points}
    \end{subfigure}
    \caption{Boxplot of model sizes for different variations of TIR and TIRMOO considering (a) every dataset and (b) the datasets selected by point.}
\end{figure}

Focusing only on the TIR and TIRMOO algorithms, we can highlight some of the key differences between these different versions of the algorithm. In Figs.~\ref{fig:size}~and~\ref{fig:size-points} we can see the boxplot of the number of nodes of the generated models for each one of these variations. The first plot is for the entire set of benchmarks while the second uses only the selection of smaller datasets. In the first plot we can see that most variants have a very similar median size, except for TIRMOO-Select with a median representing half the size of the other variants. In this same plot, TIR and TIR-points behave very similar since the penalization strategy is only active in a small selection of datasets. We can see this in effect in Fig.~\ref{fig:size-points}, in which the penalization reduces the median size to half of the TIR and TIRMOO variations.
The selection strategy adopted in TIRMOO-Select and TIRMOO-Sel-points is capable of keeping the median model size smaller than TIR-points, indicating that a good selection of the pareto front can compete with a penalization strategy in maintaining the simplicity of the models while keeping the training accuracy high.

\begin{figure}[t!]
    \centering
    \includegraphics[width=0.45\linewidth]{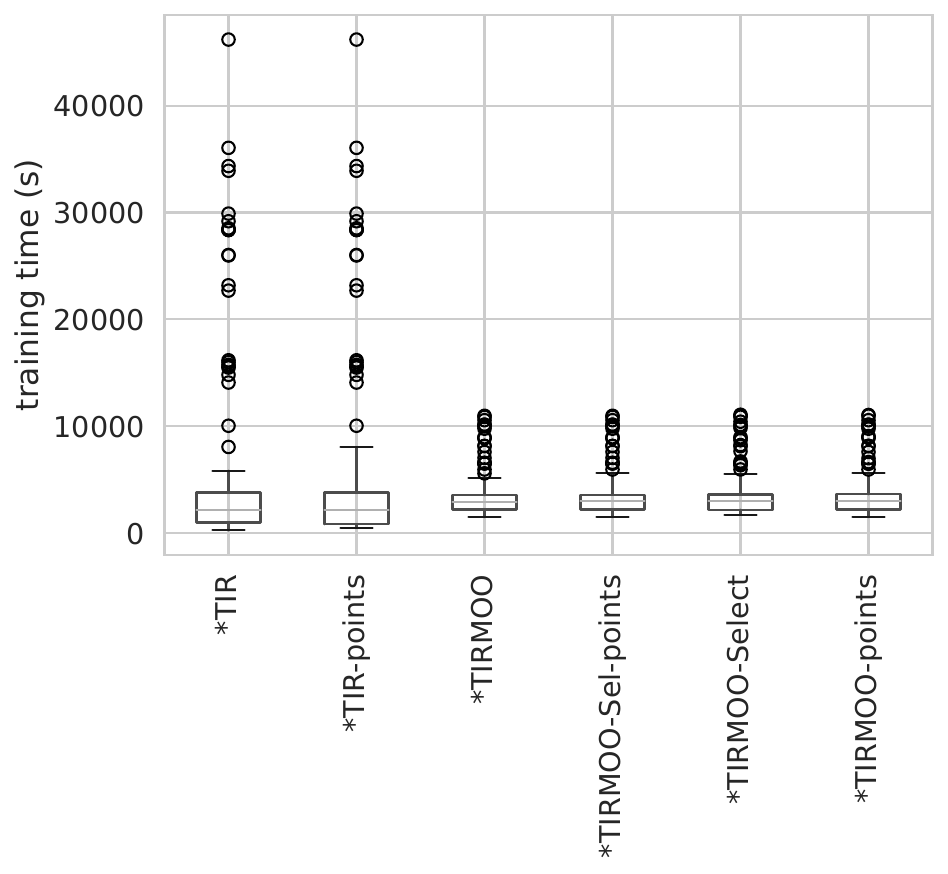}
    \caption{Boxplot of runtimes for different variations of TIR and TIRMOO. Notice that TIR considered $6$ different hyperparameters set during the gridsearch while TIRMOO only considered $2$.}
    \label{fig:time}
\end{figure}

Regarding the runtime of the different strategies, in Fig.~\ref{fig:time} we can see that the median runtime of the MOO strategies are close to the single-objective versions and with a small number of outliers. As an observation, in the MOO versions we used only $2$ sets of hyperparameters while the single-objective used $6$, so using the same size of grid search we would expect three times the runtime.

Finally, as an example of how different the models are with different strategies, Table~\ref{tab:example} shows three examples of expressions for the 687\_sleuth\_ex1605 dataset as selected by TIRMOO, TIRMOO-Select, and TIRMOO-points. These expressions are in decreasing order of model size and we can see that, in this particular case, the smaller expression generalizes better than the others. We can see that TIRMOO-Select is more moderate in favoring a smaller expression while TIRMOO-points is more aggressive, nevertheless the Select strategy is already capable of improving the results on the test set.

\begin{table}[t!]
    \centering
    \caption{Example of models returned by TIRMOO, TIRMOO-Select, and TIRMOO-points for the 687\_sleuth\_ex1605 dataset.}
    \footnotesize
    \begin{tabular}{lccc}
      \hline 
      Alg. & model & train & test \\
      \hline 
       TIRMOO  &  $\frac{576891.07 - 5.43 sin(\frac{x_4^4}{x_2 x_0^5}) - 576737.66 e^{\frac{1}{x_4^2}} - 5.83 sin(x_2^2 x_1^4)}{1 + 0.05 sin(x_3^4) + 0.06 sin(x_2^3 x_1^4)}$ & $0.87$ & $-0.43$ \\ \hline 
       TIRMOO-Select & $12111.8 - 11898.06 e^{\frac{1}{x_4}} - 7.07 \sin(x_1^4 x_2^3) - 3.52 \sin(x_3^4)$ & $0.74$ & $0.20$ \\ \hline 
       TIRMOO-points & $201.08 - 1061.77 tanh(\frac{1}{x_4})$ & $0.64$ & $0.53$ \\
       \hline 
    \end{tabular}
    \label{tab:example}
\end{table}

In Table~\ref{tab:hypervolume} we compare the hypervolumes of the fronts generated by TIRMOO and TIRMOO-points on a sample of the benchmarks. These values are averaged over the $10$ different seeds for each dataset. The values of $R^2$ are clipped between $[0, 1]$ and the values of the model size are normalized. The hypervolumes are then calculated using the reference point $[0, 1]$. Notice that TIRMOO and TIRMOO-Select share the same front, differing only on which solution is returned, thus they have the same hypervolumes and are omitted from the table. We can see from this table that the hypervolume is overall high for all of the strategies but suffers a degradation when applying a penalization into the accuracy objective-function.

\begin{table}[t!]
    \centering
    \caption{Average and standard deviation of the hypervolumes for a selection of the datasets using three different reference points for $R^2$ (maximization) and number of nodes (minimization).}
    \begin{tabular}{l|c|c}
    \hline 
    Dataset & TIRMOO & TIRMOO-points \\
    \hline 
    687\_sleuth\_ex1605 & $0.87 \pm 0.02$ & $0.71 \pm 0.04$ \\
    659\_sleuth\_ex1714 & $0.96 \pm 0.00$ & $0.90 \pm 0.02$ \\
    678\_visualizing\_environmental & $0.66 \pm 0.04$ & $0.47 \pm 0.05$ \\
    594\_fri\_c2\_100\_5 & $0.80 \pm 0.02$ & $0.74 \pm 0.02$ \\
    210\_cloud & $0.94 \pm 0.01$ & $0.90 \pm 0.02$ \\
    706\_sleuth\_case1202 & $0.87 \pm 0.02$ & $0.79 \pm 0.02$ \\
    579\_fri\_c0\_250\_5 & $0.92 \pm 0.00$ & $0.80 \pm 0.03$ \\
    613\_fri\_c3\_250\_5 & $0.92 \pm 0.01$ & $0.77 \pm 0.01$ \\
    596\_fri\_c2\_250\_5 & $0.91 \pm 0.01$ & $0.77 \pm 0.02$ \\
    601\_fri\_c1\_250\_5 & $0.92 \pm 0.01$ & $0.74 \pm 0.03$ \\
    \hline 
    \end{tabular}
    \label{tab:hypervolume}
\end{table}

\section{Conclusion}\label{sec:conclusion}

In this paper we investigated the use of multi-objective optimization with the Transformation-Interaction-Rational (TIR) evolutionary algorithm to alleviate the overfitting problem observed in small data scenarios. More specifically, we have added support to the Non-dominated Sorting algorithm for the reproduction step and Crowding Tournament selection for the parental selection. The hypothesis is that by evolving a Pareto front of solutions that minimize the approximation error and the expression size, would enforce the generation of simpler solutions with high accuracy, leading to a better extrapolation behavior.

The performance of the MOO version, called TIRMOO, was assessed using the SRBench benchmark and compared not only to its predecessors, ITEA and TIR, as well with other symbolic regression models supported by the benchmark. We have tested four different variations of this algorithm: i) a plain version of MOO, ii) the use of MOO and the penalization strategy used in the single-objective approach, iii) a MOO version where the returned model is the simplest within $95\%$ of the best accuracy and, iv) the application of the third strategy, but only for small datasets.

During the evolutionary process, the MOO approach creates a selective pressure for solutions that, given a fixed value for one objective, optimize the other one. So, among different models with similar accuracy, it will apply a selective pressure to pick the smallest one, as per the domination criteria, naturally applying the Occam's razor.

We analysed the performance of these variations using error bars of the median of medians of the results, critial difference diagrams, and histograms of frequency of high-ranks. The results indicate that, overall, the fourth strategy, called TIRMOO-Sel-points, was the most successful among the tested approaches and returned better results than the best TIR variation. Focusing on a subset of the benchmarks, the greatest difference was observed in the Friedman datasets, as TIRMOO, TIRMOO-Sel-points, TIRMOO-points presented a noticeably better result than the lower part of the rank. Regarding the small datasets, there was barely noticeable difference between the ranked algorithms, but with a slightly advantage of TIRMOO-Sel-points strategy w.r.t. the other variants. This suggests that the strategy for selecting the best solution from the Pareto front may be important to improve (even if by a small margin) the generalization accuracy. Not only that, but this strategy must take into account the uncertainty about the model and the data.

Even though the more aggressive penalization strategy seems to work better on the smaller datasets (using the number of data points time dimension as a criteria), TIRMOO-Select seems to give a good compromise between a harsh approach and relying only on the most accurate solution of the Pareto front. A broader study on these particular datasets involving the top algorithms is necessary to verify their particularities and assess whether they are inappropriate for a benchmark or if they require special treatment in the separation between training and validation sets.

For future works, we will investigate different selection strategies, such as Akaike information criterion (AIC), Bayesian information criterion (BIC), and Minimum Description Length (MDL). Also, we will test different secondary objectives for the simplicity measure, while the number of nodes is a good proxy for that, it does not take into account the linearity, smoothness and other desiderata in a regression model.

\backmatter

\bmhead{Acknowledgments}

This project is funded by Funda\c{c}\~{a}o de Amparo \`{a} Pesquisa do Estado de S\~{a}o Paulo (FAPESP), grant number 2021/12706-1 and CNPq through the grant 301596/2022-0.

\section*{Declarations}

The authors have no conflicts of interest to declare that are relevant to the content of this article.

\section*{Author contribution}

F.O.F. wrote the main manuscript, implemented the algorithm, executed the experiments, and prepared all figures and tables.

\bibliography{bibliography}

\end{document}